\documentclass{article}
\usepackage[square,numbers,sort&compress]{natbib}
\usepackage{graphicx} 
\usepackage{mathtools,amsmath}
\usepackage{amsfonts}
\usepackage{subcaption}
\usepackage{booktabs, boldline}
\usepackage{wrapfig}
\usepackage{setspace}
\usepackage{float}
\usepackage{sidecap}
\usepackage{tikz-cd} 
\newcommand{\tablepath}{tables}
\emergencystretch=\maxdimen
\usepackage[utf8]{inputenc}
\usepackage{amsmath}
\usepackage{amsfonts}
\usepackage[T1]{fontenc}
\usepackage{charter}
\sidecaptionvpos{figure}{t}
\let\oldabstract\abstract
\let\oldendabstract\endabstract
\makeatletter
\renewenvironment{abstract}
{%
               {\list{}{\addtolength{\leftmargin}{2em} 
                        \listparindent 1.5em%
                        \itemindent    \listparindent%
                        \rightmargin   \leftmargin%
                        \parsep        \z@ \@plus\p@}%
                \item\relax}%
               {\endlist}%
\oldabstract}
{\oldendabstract}
\makeatother

\usepackage[colorlinks = true,
            linkcolor = blue,
            urlcolor  = blue,
            citecolor = blue,
            anchorcolor = blue]{hyperref}

\definecolor{darkred}{rgb}{0.7, 0.0, 0.0}
\usepackage[lmargin=2cm,rmargin=2cm,tmargin=2cm,bmargin=2cm]{geometry}
\title{Multi-Lattice Sampling of Quantum Field Theories\\ via Neural Operator-based Flows}
\author{
    Bálint Máté
    \and François Fleuret
    }

\date{ University of Geneva \\ \vspace{6pt} \texttt{\{balint.mate,francois.fleuret\}@unige.ch}}
\begin{document}
\maketitle
\begin{abstract}
    {
    We consider the problem of sampling lattice field configurations on a lattice from the Boltzmann distribution corresponding to some action.
    Since such densities arise as  approximationw of an underlying functional density, we frame the task as an instance of operator learning. We propose to approximate a time-dependent neural operator whose time integral provides a mapping between the functional distributions of the free and target theories.
    Once a particular lattice is chosen, the neural operator  can be discretized to a finite-dimensional, time-dependent vector field which in turn induces a continuous normalizing flow between finite dimensional distributions over the chosen lattice. This flow can then be trained to be a diffeormorphism between the discretized free and target theories on the chosen lattice, and, by construction, can be evaluated on different discretizations of spacetime. We experimentally validate the proposal on the 2-dimensional $\phi^4$-theory to explore to what extent such operator-based flow architectures generalize to lattice sizes they were not trained on, and show that pretraining on smaller lattices can lead to a speedup over training directly on the target lattice size.}
\end{abstract}
\section{Introduction}
Consider an action $\mathcal{S}$, characterizing a quantum field theory, and $S$, a discretized representation  of  $\mathcal{S}$ on a lattice. \citet{albergo2019flow} suggest a method for sampling from the lattice quantum field theory described by $S$ by using a normalizing flow parameterizing a probability density $q_\theta(\phi)$ of discrete fields over the lattice, and optimizing the parameters $\theta$ until $q_\theta(\phi)$ closely approximates $\frac{e^{-S[\phi]}}{Z}$, where $Z=\int [d\phi] e^{-S[\phi]}$ is the normalizing constant.
\vspace{8pt}

Operator learning  promotes the viewpoint that the lattice/mesh is merely a computational tool, and the model should capture the underlying continuous physics. \citet{kovachki2021neural}  refer to this property as discretization invariance. \footnote{Discretization invariance means that the neural operator evaluated on finer and finer discretizations approximates the continuous operator. Thus, strictly speaking, it is not a requirement of invariance rather one of convergence.} In this work, we apply the same idea to the task of sampling from lattice quantum field theories, motivated by the fact that lattice field theories also emerge as the discretization of continuous field theories. 
\vspace{8pt}

{
\subsection*{(Lattice) Quantum  Field Theory}
A quantum field theory is defined by an action functional $\mathcal{S}$ that assigns a scalar to field configurations on some domain, typically denoted by $\phi(x)$, where $ x $ represents spacetime coordinates. After a Wick rotation (a transformation that makes time act like a spatial dimension), the problem of interest is to sample from the functional density $\frac{1}{\mathcal{Z}} e^{-\mathcal{S}[\phi]}$, where $\mathcal{Z}$ is a normalizing constant, often called the partition function. This density represents the probability distribution over all possible field configurations in Euclidean spacetime. The goal is to understand the behavior of the system by computing expected values of observables (like correlation functions) with respect to $\frac{1}{\mathcal{Z}} e^{-\mathcal{S}[\phi]}$.

\vspace{10pt}

In quantum field theory, an \textit{observable}  $\mathcal O$, is simply a function of the field configuration $\phi(x)$. For example, observables could include the average field value, or correlations between field values at different points in space. When working with the functional density $\frac{1}{\mathcal{Z}} e^{-\mathcal{S}[\phi]}$, the objective is to compute the \textit{expected values} of these observables with respect to this distribution, i.e. quantities of the form

\begin{equation}
    \langle O \rangle = \int \mathcal{D} \phi \, O[\phi] \, \frac{1}{\mathcal{Z}} e^{-\mathcal{S}[\phi]},
\end{equation}

This integral represents the average of $O$ across all possible field configurations, weighted by the probability density $\frac{1}{\mathcal{Z}} e^{-\mathcal{S}[\phi]}$. These expected values correspond to measurable, physical quantities in experiments, as they represent the statistical average over many possible configurations of the quantum field. In practice, we cannot measure individual field configurations; instead, we observe properties like particle masses, energy densities, and correlation functions, each of which can be computed as the expected value of an appropriate observable. Therefore, expected values provide the essential link between theoretical calculations and physical measurements, making them central to understanding the field's behavior.

\vspace{10pt}

To do this computationally, quantum field theories are typically \textit{discretized} by approximating the continuous spacetime domain with a lattice structure. This \textit{discretization} translates the continuous field $\phi(x)$ into a set of field values $\phi_i$ defined at specific lattice points. By introducing a finite lattice spacing $a$, we can approximate the continuous integrals over spacetime by discrete sums over lattice points, making the model tractable for numerical simulations. However, discretization inherently introduces approximations. For instance, the lattice spacing $a$ imposes a high-frequency cutoff, as fluctuations with wavelengths smaller than $a$ (frequencies above $1/a$) cannot be captured on the lattice. Additionally, the lattice domain is typically chosen to be \textit{periodic} to reduce edge effects, effectively wrapping around in each dimension. The size of this periodic domain must be sufficiently large to avoid \textit{finite-size effects}, where the limited domain size influences physical quantities, potentially distorting results.

\vspace{10pt}

While discretization makes the theory computationally feasible, the ultimate goal is to recover the properties of the original continuous quantum field theory. To achieve this, one must take the \textit{continuum limit}, in which the lattice spacing $ a $ is taken to zero. In this limit, the lattice approximation is refined such that finer and finer details of the field configurations are captured, and high-frequency fluctuations, initially filtered out by the discrete lattice, are restored. As $ a \to 0 $, the discretized sums over lattice points approach the continuous integrals that define the original theory, and the lattice version of the expected values $\langle O \rangle$ should converge to their continuous counterparts. In practice, this requires computing the expected values of observables at several different lattice spacings and then \textit{extrapolating} to estimate their values as $a \to 0$. This extrapolation process allows us to obtain results that more accurately reflect the true, measurable quantities in the continuous theory. The continuum limit is essential because only in this limit do the results accurately reflect the underlying physics without artifacts from discretization. However, taking this limit is computationally demanding, as it requires not only reducing $a$ but also increasing the number of lattice points to maintain the physical size of the domain, thus ensuring that finite-size effects remain negligible.

\vspace{10pt}

In this work, we address this last aspect, the issue of taking the continuum limit. Traditionally, this process requires computing observables at various lattice spacings and then extrapolating to $a \to 0$. We propose an alternative approach by training a single model that can be used across different resolutions. This model leverages the concept of neural operators, which are designed to learn  continuous operators between function spaces.
This approach has the potential to simplify the extrapolation process, as the trained model can provide expected values for any lattice spacing, allowing for a more continuous mapping of observables as the continuum limit is approached. Consequently, we can obtain a more accurate representation of the physical quantities in the continuum theory while mitigating the computational burden typically associated with conventional, single-lattice methods.

\vspace{10pt}

It is important to note that this approach is related to the \textit{renormalization group} (RG) methods, which also seek to understand how physical quantities change with scale. However, rather than renormalizing the action, we focus on discretizing the action at different resolutions and aim to learn a continuous representation of the underlying field theory. Our goal is to create a model that captures the essential features of the quantum field across different scales, thus enabling a more direct pathway to the continuum limit.
}

{Suppose now that the field theory is defined on some domain $D$, i.e. fields are functions living on $D$. We can reduce this system to a finite dimensional one by introducing a lattice, i.e. a collection of regularly spaced points, in $D$, and model fields as functions on this finite collection of lattice sites.} In this setting, a continuous normalizing flow \cite{chen2018neural} can be introduced as a time-dependent vector field $V_t$ that parametrizes the direction along which probability mass moves. Generalizing this idea, we propose to parametrize a time-dependent operator $\mathcal V_t$ from the space of functions on $D$ to itself that defines the direction in which functional probability mass moves. Such  an operator can then be used to map the functional distributions $[\mathcal D\phi(x)]\mathcal Z_0^{-1}e^{-\mathcal S_0[\phi(x)]}, [\mathcal D\phi(x)]\mathcal Z_1^{-1}e^{-\mathcal S_1[\phi(x)]}$, {with corresponding actions $\mathcal S_0,\mathcal S_1$,} to one another. Computationally the operator $\mathcal V_t$ can only be accessed by a choice of a lattice which induces a vector field $V_t$ as the discretization of  $\mathcal V_t$. We then train this vector field to be a diffeomorphism between the discretized free and target theories, $[d\phi] Z_0^{-1} e^{-S_{0}[\phi]}$ and $[d\phi] Z_1^{-1}e^{-S_1[\phi]}$.
The upside of using an operator-based flow will be that a single model can be used to operate on multiple discretizations of the same underlying continuous system. Figure \ref{fig:tikz} provides a schematic overview of the objects and their relation in this paragraph.

\begin{center}
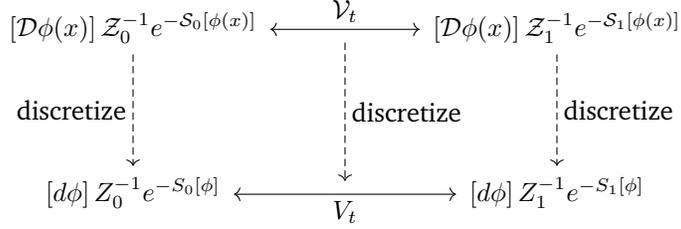

\begin{tikzcd}[column sep=2cm,row sep=1.5cm,every label/.append
    style={font=\normalsize}]
    \left[\mathcal D\phi(x)\right]\mathcal Z_0^{-1}e^{-\mathcal S_0\left[\phi(x)\right]}
    \arrow[leftrightarrow,""{name=V1}]{r}{\mathcal V_t}
    \arrow[swap,dashrightarrow]{d}{\text{discretize}} 
    & \left[\mathcal D\phi(x)\right]\mathcal Z_1^{-1}e^{-\mathcal S_1\left[\phi(x)\right]}
    \arrow[dashrightarrow]{d}{\text{discretize}}  \\

    \left[d\phi\right]Z_0^{-1} e^{-S_{0}\left[\phi\right]} 
    \arrow[swap,font=\Huge,leftrightarrow,""{name=V2}]{r}{\Large V_t}
    \ar[dashrightarrow,from=V1, to=V2, "\text{discretize}", shorten <= 5pt, shorten >= 5pt]
    & \left[d\phi\right]Z_1^{-1} e^{-S_{1}\left[\phi\right]}
\end{tikzcd}

\captionof{figure}{Schematic overview of the probability distributions of interest. The top row shows the functional distributions of the free theory and the target theory connected by the time dependent operator $\mathcal V_t$. Moving to the bottom row corresponds to approximating each object from the top row on a discrete lattice. In particular, in the bottom row all objects are finite dimensional, well-defined and can be worked with numerically.}
\label{fig:tikz}
\end{center}

{

\subsection*{Related Work}
\citet{albergo2019flow} proposed the use of normalizing flows for sampling from lattice field theories. Since then, a variety of methods have emerged for sampling scalar \cite{nicoli2021estimation,hackett2021flow,nicoli2021machine,gerdes2022learning,caselle2022stochastic,singha2022conditional,albandea2022learning,pawlowski2022flow,nicoli2023detecting} as well as gauge field theories \cite{flowsforlattice2,finkenrath2022tackling,flowsforlattice3,PhysRevLett.128.032003,bacchio2022learning,albergo2021flow,abbott2022gauge}. As the action of the theory is available, one can train these normalizing flow models in a data-free setting, using only evaluations of the action and its gradients as the training signal. The simplest way of doing this is to rely on the reverse Kullback–Leibler (KL) divergence, $KL(q_\theta, p)$, between the density represented by the flow $q_\theta$ and the target Boltzmann density $p(\phi)=e^{-S(\phi)}/Z$. This approach encourages the learned density $q_\theta$ to match the high-probability regions of the target distribution, which can be an effective strategy for capturing prominent features of $p$.
However, minimizing the reverse KL divergence can lead to mode collapse {\cite{nicoli2023detecting,vaitl2022gradients,vaitl2022path,vaitl2024fast,mate2023learning}}, meaning that the model density $q_\theta$ fails to represent the full diversity of the target distribution by focusing too narrowly on dominant modes. This limitation has motivated recent work to explore alternative training objectives that mitigate mode collapse. \citet{vaitl2022gradients,vaitl2022path,vaitl2024fast} introduce improved estimators of the  KL divergence. A different objective motivated by the continuity equation describing the flow of probability mass between the target and prior was considered in  \cite{mate2023learning}.

Most of these approaches focus primarily on training models for a single lattice size, which limits their adaptability across different lattice spacings. Notably, \citet{gerdes2022learning} devise a method to embed a discrete convolutional kernel into a kernel for a larger lattice. In contrast, our approach treats the kernels as inherently continuous objects from the outset, enabling us to work with their discretizations across multiple lattice sizes.
}

\section{Background}
\label{sec:background}
\subsection{Continuous Normalizing Flows}
\label{sec:flows}
A continuous normalizing flow \cite{chen2018neural} is a density estimator {and sampler} that operates by pushing forward a simple, usually Gaussian, initial density $q_0$ along a parametric, time-dependent vector field $V_\theta: [0,1]\times \mathbb{R}^n \rightarrow  \mathbb{R}^n$. Explicitly, the pushforward density $q_\theta$ is given by 
\begin{align}
    \label{eq:flow}
    \log q_\theta(x_1) = \log q_0(x_0) + \int_1^0dt\, \nabla\cdot V_\theta(t,x_t)
\end{align}
where $\nabla$ is the divergence operator in the spatial coordinates and $x_t$ is the integral curve of $V_\theta$ that passes through $x_1$ at $t=1$.
In this work, all normalizing flows will be continuous, and we will refer to them as normalizing flows or even just flows for brevity.
\paragraph{Boltzmann distributions} The Boltzmann distribution of an energy function\footnote{Assuming that $ \exp(-S)$ is integrable.}  $S:\mathbb{R}^n \rightarrow\mathbb{R}$ is a probability distribution with density 
\begin{equation}
    p(x) = \frac{1}{Z}e^{-S(x)}
\end{equation}
where $Z=\int dx\,e^{-S(x)}$ is the normalizing constant ensuring that the density function integrates to $1$.
Boltzmann distributions appear in the context of the canonical ensemble, a statistical ensemble that describes a system in thermal equilibrium with an external heat reservoir. Such Boltzmann distributions describe the molecular systems in thermal equilibrium as well as  Wick-rotated quantum field theories. Learning to sample from Boltzmann distributions using only the energy function (i.e. without true samples) can be done by training a normalizing flow, usually, to minimize the reverse KL divergence

\begin{align}
    KL(q_\theta,p) &= \mathbb{E}_{x\sim q_\theta}\left[\log q_\theta(x)-\log p(x)\right]\\&=\mathbb{E}_{x\sim q_\theta}\left[\log q_\theta(x)+f(x)\right] + \log Z
\end{align}
where $q_\theta$ is the density realized by the normalizing flow (\ref{eq:flow}).  Once a density $q_\theta$, approximating $p=Z^{-1} e^{-f}$, is learnt, one can use importance sampling to correct for small inaccuracies of $q_\theta$ when  estimating the expected value of an observable $\mathcal O$, {i.e. a real-valued function of the configurations,}
\begin{equation}
    \label{eq:reweighting}
\langle \mathcal O \rangle := \mathbb{E}_{\phi\sim p}\left[\mathcal O (\phi)\right] = \mathbb{E}_{\phi\sim q_\theta}\left[\mathcal O (\phi) \frac{p(\phi)}{q_\theta(\phi)} \right]
\end{equation}

\subsection{The $\phi^4$ (Lattice) Quantum Field Theory\protect \footnote{We recommend the book \citep[Chapter 15]{thijssen_2007} for further details on lattice field theories.}}
Let us now consider the Euclidean action on real-valued, {continuous} scalar fields $\phi(x)$ with periodic boundary conditions on the $D$-dimensional  hypercube of edge length $L$, $\phi:(\mathbb{R}/L\mathbb{Z})^D\rightarrow \mathbb{R}$, for some constants $m^2$ and $g$
\begin{align}
    \label{cont_action}
    \mathcal S[\phi] &= \int_{(\mathbb{R}/L\mathbb{Z})^D} d^D x \left[(\nabla\phi(x))^2+m^2\phi(x)^2+ g\phi(x)^4 \right]
\end{align}
{For the rest of this work, we will drop the the argument of $\phi(x)$ and denote fields by $\phi$ to unclutter notation.}
To estimate the expectation value of an observable $\mathcal O$, we need to average over all field configurations that satisfy the boundary conditions, with each configuration weighted by  its Boltzmann weight
\begin{equation}
    \label{eq:path_integral}
    \langle \mathcal O \rangle  = \frac{\int \mathcal D \phi \,\mathcal O [\phi]e^{-\mathcal S[\phi]}}{\int \mathcal D \phi\, e^{-\mathcal S[\phi]}}
\end{equation}
\vspace{8pt}

Equations (\ref{cont_action}) and (\ref{eq:path_integral}) describe an infinite dimensional system, where microstates are functions on $(\mathbb{R}/L\mathbb{Z})^D$. To tackle it numerically, one first needs to discretize the domain $(\mathbb{R}/L\mathbb{Z})^D$, the action $\mathcal S$, and fields $\phi(x)$ to a lattice. This comes at the cost of losing the information contained in the high-frequency components as the highest possible frequency of a periodic function on a lattice with edge length $L$ with $N$ nodes is $\frac{2\pi N}{L}$. The hope is that one can do the same on larger and larger lattices, and as the lattice approaches the continuum limit, the error due to discretization converges to zero.
\subsubsection*{Discrete representations on lattices}
To discretise the action, we consider fields living on the points located at  $\left\{\tfrac{0}{N},\tfrac{L}{N},...,\tfrac{(N-1)L}{N}\right\}^D$ forming a periodic lattice with cardinality $N^D$ and lattice spacing $a=L/N$. We then turn integrals into sums and differentials into differences between nearest neighbors 
\begin{align}
    \partial_i \phi &\rightarrow \frac{1}{a} \phi(x+\mu_i)-\phi(x)\\
    \int_{(\mathbb{R}/L\mathbb{Z})^D} d^D x &\rightarrow a^{D}\sum_{x}
\end{align}
where $\mu_i$ denotes the generator of the lattice along the $i-$th coordinate axis, {i.e. the displacement vector along neighboring lattice sites}. After these substitutions we end up with the following discretised action on the lattice,
\begin{equation}
    \label{eq:disc_action}
    S[\phi] = a^{D}\left\{\frac{1}{a^2}\sum_{x,\mu} (\phi_{x+\mu} -\phi_x)^2+ \sum_x {m^2}\phi_x^2+ g\phi_x^4\right\}
\end{equation}
where $x$ runs over the lattice sites and $\mu$  over the generators of the lattice. 
It is customary to absorb all the occurrences of $a$ in the above formula by rescaling $\phi$ 

\begin{equation}
    \phi \rightarrow a^{D/2-1}\phi, \qquad m \rightarrow am, \qquad g \rightarrow a^{4-D}g
\end{equation}    
This results in an alternative form of the action
\begin{equation}
    \label{eq:disc_action_nono}
    S[\phi] =\sum_{x,\mu} (\phi_{x+\mu} -\phi_x)^2+ \sum_x { m^2}\phi_x^2+  g\phi_x^4
\end{equation}
While the action (\ref{eq:disc_action_nono}) has the advantage of not being dependent on the lattice spacing $a$, we will continue working with (\ref{eq:disc_action}) keeping the relation between different lattice sizes and  to the underlying continuous setting explicit.

\subsection{Neural Operators}
\label{sec:neural_operator}
\begin{figure}[H]
    \centering
    \includegraphics[width=\textwidth,trim={0cm .2cm 0cm 2cm},clip]{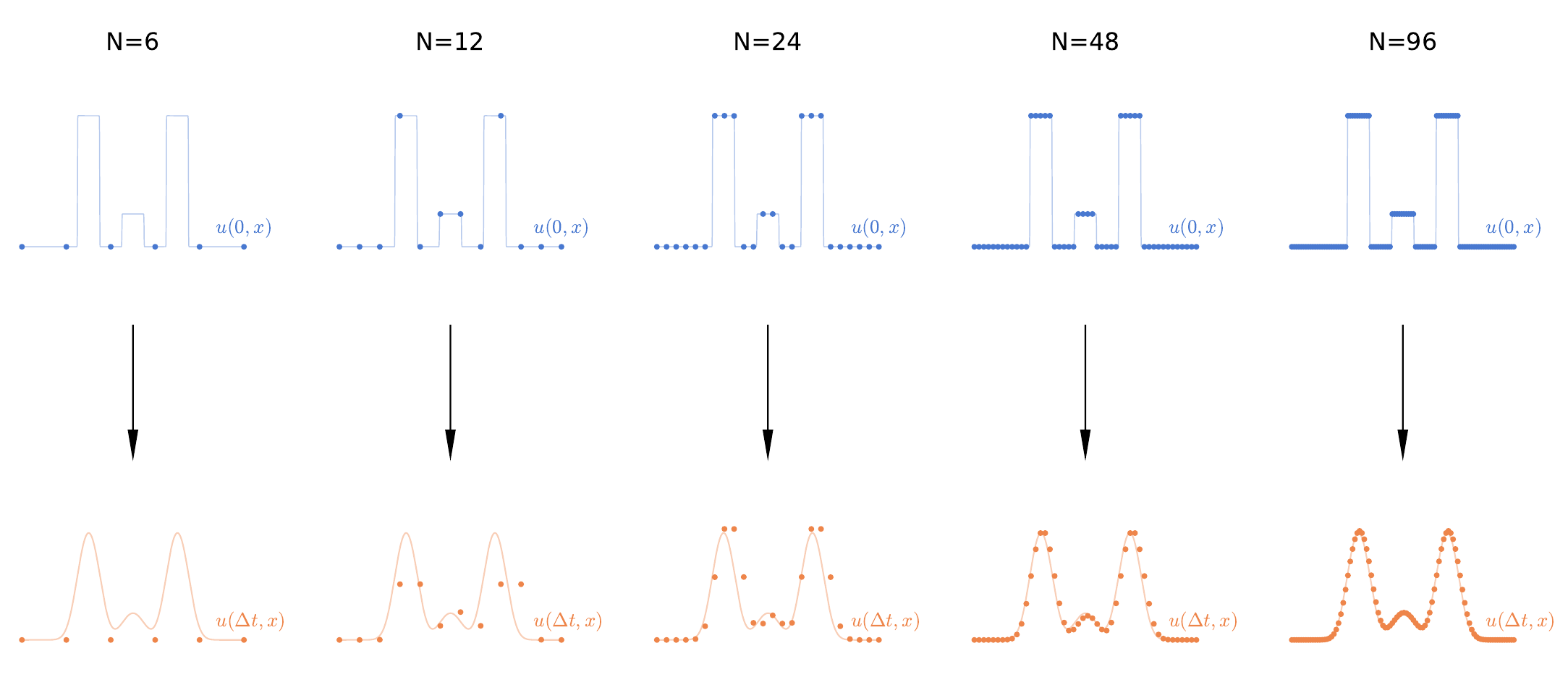}
    \caption{An operator that maps an initial condition $u(0,x)$ (top row) to its time-evolved state $u(\Delta t,x)$ (bottom row), where the time evolution is  given by the heat equation $\Delta u  = \partial_t u $. The blue dots denote the evaluation of $u(0,x)$ on a discrete mesh, while the orange dots denote the output of the operator (a convolution in this case) evaluated on that same mesh. As the mesh gets denser, the operator becomes a better approximation of the map between the continuous $u(0,x)$ (blue curve) and $u(\Delta t,x)$ (orange curve). {For our application, we will be interested in the time evolution of probability density of fields (corresponding to $u$ in the plots), along the time interval $[0,1]$ connecting the free theory to a interacting theory.}}
    \label{fig:operators}
\end{figure}

Neural Operators \cite{kovachki2021neural} are trainable function-to-function mappings, both their domains and codomains are infinite dimensional function spaces. In practice, one works with neural operators by choosing a mesh/lattice $ X \subset \mathbb{R}^n$, representing functions by their evaluations on $X$ and let the neural operator operate on this  collection of  evaluations. By design, neural operators can be evaluated on lattices of different size. Importantly, if a neural operator is applied to a sequence of meshes $X_i$, approaching the continuum limit $X_i \rightarrow \mathbb{R}^n$, it converges to the underlying continuous operator. The main use case of neural operators is to approximate the solution of partial differential equations, i.e. learn the mapping from an initial condition to the time evolved state after some time $\Delta t$ (Figure \ref{fig:operators}), but have also been applied for multi-resolution generative modelling \cite{voleti2021multi,hagemann2023multilevel}. We will use them for parametrizing a flow, i.e. a vector field $\mathcal V_t$ connecting the free theory (base density) to the $\phi^4$-theory (target density) in a way that can be evaluated at any mesh. 

\section{Flows parametrised by neural operators}
\label{sec:our_arch}
When designing the architecture we kept the following considerations in mind
\begin{enumerate}
    \setlength\itemsep{.01em}
    \item The architecture should be a neural operator as described in Section \S \ref{sec:neural_operator}.
    \item The architecture should respect the symmetries of the target density.
    \item The architecture should be such that its divergence is reasonably cheap to compute, since it will be integrated over trajectories to compute the density represented by the flow (Equation \ref{eq:flow}).
\end{enumerate}
\paragraph{The architecture in a nutshell}
The output of the {architecture} is given by convolving the input with a parametric continuous kernel $ K_\theta:(\mathbb{R}/L\mathbb{Z})^D\rightarrow \mathbb{R}$. This guarantees that the first requirement is satisfied, and the second one forces $ K_\theta$ to be spherically symmetric.
Regarding the last one, we take inspiration from \citet{chen2019neural}, and use the combination of a conditioner function $h_i =c(\phi_{-i})$, whose output at any coordinate is independent of the same coordinate of the input, and a transformer function $f_i = \tau(h_i,\phi_i)$ that combines the conditioning and the input. The advantage of this architecture is that its divergence is  $\sum_i(\partial_2 \tau)$ and thus the capacity of $c$ can be cheaply increased.

\begin{figure}[H]
    \centering
    \includegraphics[width=.9\textwidth,trim={6cm .3cm 0.5cm .2cm},clip]{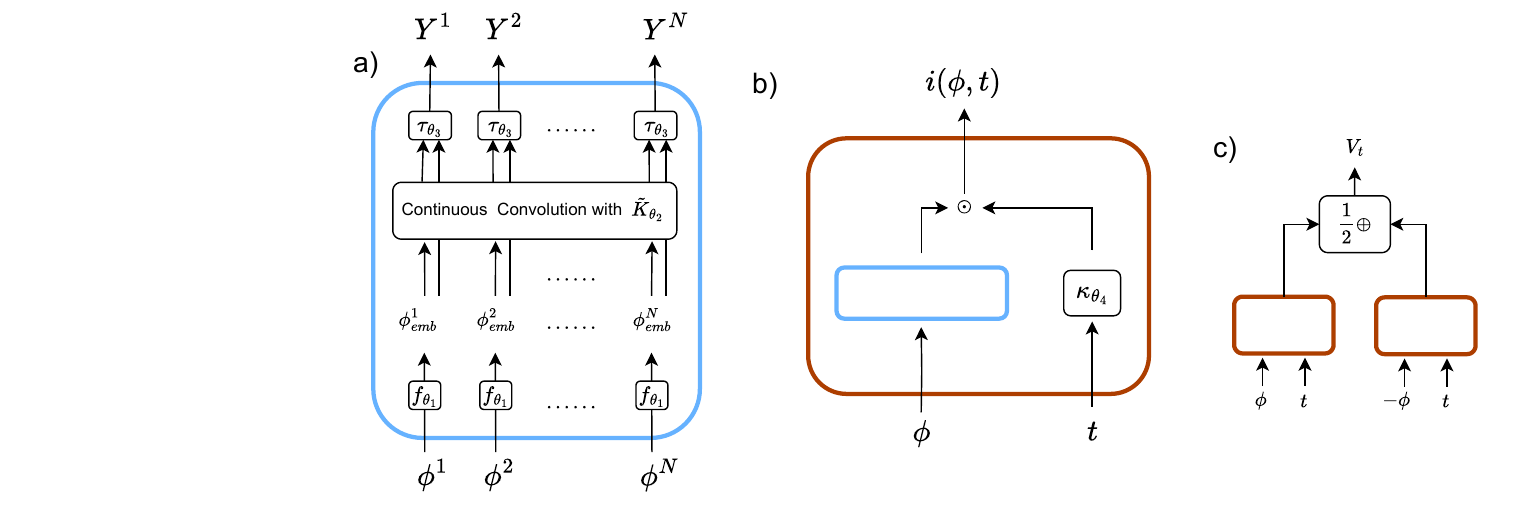}
    \caption{Sketch of the architecture. (a) Pointwise field embedding via $f_{\theta_1}$, continuous convolution with kernel $\tilde K_{\theta_2}$ to aggregate local information, and combination of the pointwise field values with neighborhood information via $\tau_{\theta_3}$. (b) Contracting the channels with learnable time-dependent weights given by $\kappa_{\theta_4}(t)$. (c) Averaging of the preceding steps over the sign of $\phi$ to enforce the $\mathbb Z_2$-symmetry of the theory.}
    \label{fig:architecture}
\end{figure}

\paragraph*{The architecture in detail}
Let now $\phi \in \mathbb{R}^{N\times .... \times N}$ be a discretized scalar field on a lattice. The architecture then consists of the following sequence of steps, where the subscript $\theta_i$ denotes trainable parameters,
\begin{enumerate}
    \setlength\itemsep{.01em}
    \item Use a per-node neural network $f_{\theta_1}$ to embed the field values, $\phi_{emb} = f_{\theta_1}(\phi) \in \mathbb{R}^{c\times N\times .... \times N}$, where $c$ is the number of channels.
    \item Use a neural network to parametrize $c$-many continuous spherically symmetric kernels $K_{\theta_2}(r)$. Let then $\tilde K_{\theta_2}$ be the evaluation of the continuous kernels on the lattice.
    \item Mask out the origin of the discrete kernel, i.e. set $\tilde K_{\theta_2}[:,\textbf 0] = 0$. {This is done to avoid the dependence of every coordinate of the output on the same coordinate of the input and thus provide an efficient computation of the divergence \cite{chen2019neural}.}
    \item Perform a the channel-wise convolution $\phi_{emb} \star \tilde K_{\theta_2}$ and denote the result by $C\in \mathbb{R}^{c\times N\times .... \times N} $. Because of the previous step, $C_i$ is independent of $\phi_i$, and we will call it the conditioner \cite{chen2019neural}.
    \item Apply a per-node neural network $\tau_{\theta_3}$ to the concatenation $(C,\phi_{emb})$ with output $Y=\tau_{\theta_3}(C,\phi_{emb}) \in  \mathbb{R}^{T\times N \times .... \times N}$.
    \item Contract the first dimension of $Y$ with a vector of length $T$ that is the output of a neural network $\kappa_{\theta_4}$, taking time $t$ as the only input.
    \item Finally, denoting all the above steps as $i$, we set the output of the {architecture} to be $V(\phi,t) = \tfrac{1}{2}*(i(\phi,t)- i(-\phi,t))$. This enforces the $\mathbb{Z}_2$ symmetry of the system.
\end{enumerate}
To compute the divergence of the architecture one needs the Jacobians of the per-point operations $f_{{\theta_1}}$ and $K_{\theta_2}$ does not have to be differentiated through. Figure \ref{fig:architecture} shows a sketch of the architecture.

\subsubsection*{The free theory as an initial density}
The normalizing flow architecture described in Section \S \ref{sec:flows} requires an initial density from which samples can easily be drawn. Instead of sampling from a standard gaussian at every node, we choose a more physical initial density by setting $g = 0$ in the action (\ref{eq:disc_action}). This results in the free theory with a gaussian Boltzmann density that becomes diagonal in momentum space. Position and momentum space are related by a discrete Fourier transform
\begin{align}
    \label{eq:dft}
     \phi_x &= \frac{1}{\sqrt{N^D}}\sum_p \tilde\phi_p e^{i2\pi\langle p,x\rangle}\\
      \tilde\phi_p &=  \frac{1}{\sqrt{N^D}}\sum_x \phi_x e^{-i2\pi\langle p,x\rangle}
\end{align}
where $p$ runs over $\left\{-\frac{\lfloor{N/2}\rfloor}{L},....,\frac{0}{L},...,\frac{\lceil{(N-1)/2}\rceil}{L}\right\}^D$ and the prefactor $\frac{1}{\sqrt{N^D}}$ makes the map $\{\phi_x\} \leftrightarrow \{\tilde \phi_p\}$ unitary. {Since the Boltzmann distribution of the free theory is a gaussian, we need to understand the structure of its covariance matrix to generate samples from it.} This
covariance matrix is diagonalized in the momentum basis $\tfrac{1}{\sqrt{N^D}} e^{{2\pi i \langle x,p\rangle}}$ with eigenvalues 
\begin{align}
    S\left[\frac{1}{\sqrt{N^D}} e^{{2\pi i \langle x,p\rangle}} \right]=a^D\left(m^2+\frac{1}{a^2}\sum_\mu 2-2\cos(2\pi p_\mu a)\right)
\end{align}
We can thus sample configurations by  sampling the components independently in momentum space, and transform the samples back to position space with an inverse Fourier-transform. Moreover, we need to constrain the sampling to real valued fields in position space, therefore the sampling in momentum space must be constrained to the hermitian symmetric subspace of \emph{real dimension} $N^D$ of the full momentum-space of \emph{complex dimension } $N^D$.

\section{Experiments}
\label{sec:experiments}
\subsection{Multi-lattice sampling in $D=1$ dimension}
\label{sec:exp1}
\begin{figure}[H]
    \centering
    \begin{subfigure}[t]{0.31\textwidth}
        \centering
        \includegraphics[width=\textwidth]{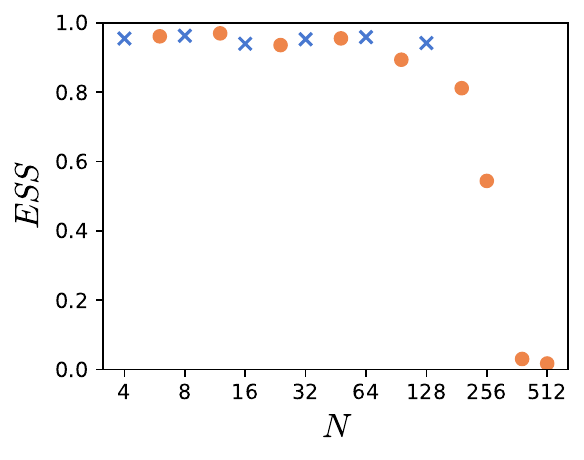}
    \end{subfigure}
    \begin{subfigure}[t]{0.32\textwidth}
        \centering
        \includegraphics[width=\textwidth]{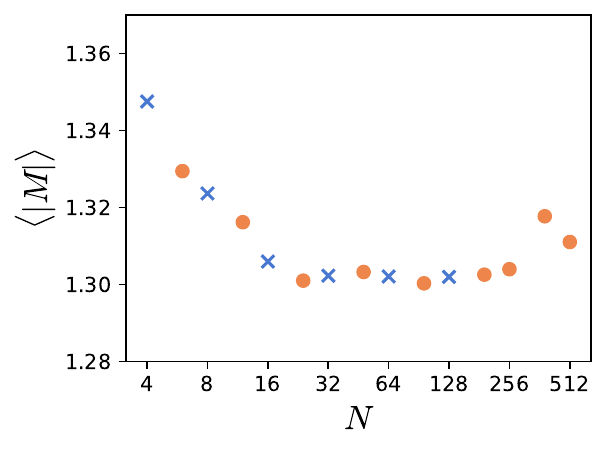}
    \end{subfigure}
    \begin{subfigure}[t]{0.33\textwidth}
        \centering
        \includegraphics[width=\textwidth]{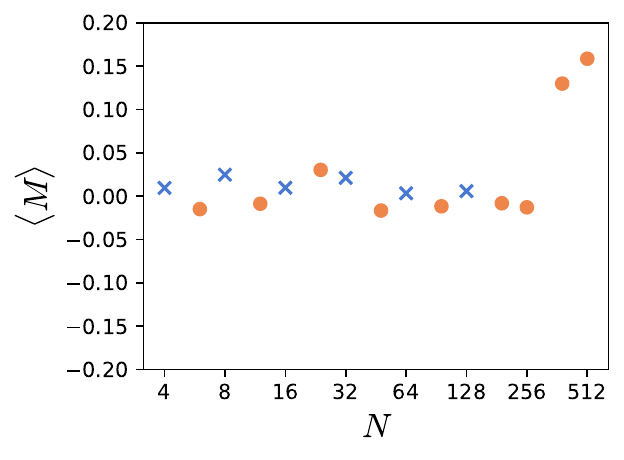}
    \end{subfigure}
    \caption{Particle in double well potential (\S \ref{sec:exp1}). $ESS, \langle M \rangle,  \langle |M| \rangle $ computed from $16384$ samples at different lattice sizes. The blue crosses correspond to lattice sizes that the model was trained on, while orange dots denote lattice sizes unseen by the network during training. {Note that the absolute magnetization converges to a value of $1.30$ as the lattice size is increased to $16$ and stabilizes at this value at larger lattice sizes.}}
    \label{fig:exp1_ess_M}
\end{figure}

\begin{figure}[H]
    \centering
    \begin{subfigure}[t]{0.5\textwidth}
        \centering
        \includegraphics[width=.9\textwidth,trim={0 .4cm 0 .2cm},clip]{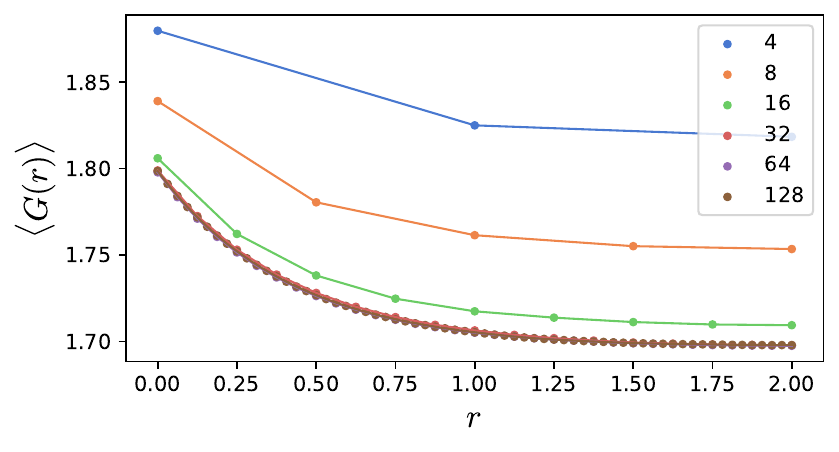}
    \end{subfigure}%
    \begin{subfigure}[t]{0.5\textwidth}
        \centering
        \includegraphics[width=.9\textwidth,trim={0 .4cm 0 .2cm},clip]{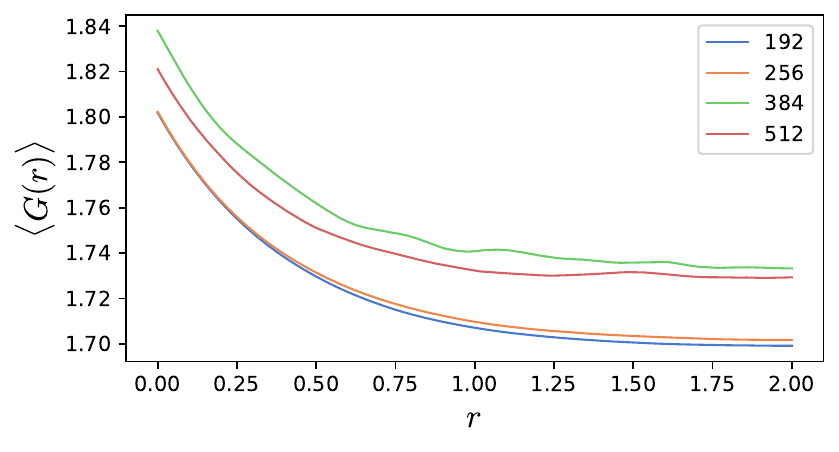}
    \end{subfigure}
    \caption{Particle in double well potential (\S \ref{sec:exp1}). The two-point correlation function $G(x,y)$  computed from $16384$ samples on lattices the model was trained on (left) and on lattices the model was not trained on (right). Because of the symmetries of the task the correlation function only depends on the distance $r = |x-y|$, thus the function $G(r)$ is plotted. {Similarly to the absolute value of the magnetization, $G(r)$ approaches the continuum limit as the model is evaluated at increasing lattice sizes.}}
    \label{fig:exp1_corr_fn}
\end{figure}

\begin{figure}[H]
    \centering
    \includegraphics[width=.9\textwidth,trim={0 .3cm 0 .2cm},clip]{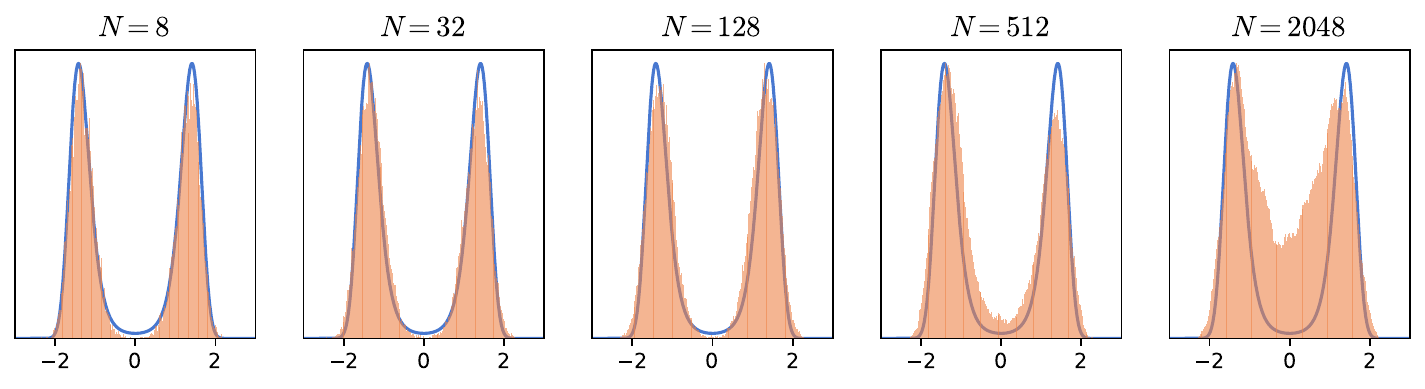}
    \caption{Particle in double well potential (\S \ref{sec:exp1}). Flattened samples (orange histogram) of the model at different lattice sizes $N=L/a$ compared to the one-dimensional Boltzmann density $e^{-m^2\phi^2 -g\phi^4}$ (blue curve).}
    \label{fig:exp1_marginals}
\end{figure}

We now work in $D=1$ dimensions. Strictly speaking, a one dimensional lattice does not correspond to a quantum field theory, rather it describes the trajectory of a quantum mechanical particle in a potential \cite{vaitl2022gradients}. Nonetheless, it's the simplest setup in which we can experiment and serves as a good starting point. We also fix $L =4, m^2=-4,g=1$ and train a single model for $5000$ steps with mesh size uniformly sampled at each training step from $N=L/a \in \{4,8,16,...,128\}$. 
We then evaluate performance on lattices of size $N=L/a$ up to 512 by sampling from the trained model to calculate the effective sample size (Figure \ref{fig:exp1_ess_M})
\begin{equation}
    ESS = \frac{\left(\tfrac{1}{N} \sum_i w_i\right)^2} {\tfrac{1}{N} \sum_i w_i^2 }
\end{equation}
where $w_i$ is the importance weight $p(\phi_i)/q_\theta(\phi_i)$.
Moreover, we estimate the expectation values of the magnetization and its absolute value (Figure \ref{fig:exp1_ess_M}) 
\begin{align} 
    M[\phi] &:=  \frac{1}{N^D}\sum_{x} \phi(x) \qquad |M|[\phi] :=  \frac{1}{N^D}\left|\sum_{x} \phi(x) \right|
\end{align}
and of the two-point correlation function (Figure \ref{fig:exp1_corr_fn}) 
\begin{equation} 
    G(x,y)[\phi] :=  \phi(x)\phi(y).
\end{equation}
 We also compare flattened samples from the model against the one-dimensional Boltzmann density of the potential $m^2 \phi^2 + g \phi^4$  (Figure \ref{fig:exp1_marginals}). {All reported observables have been reweighted from the model to the target density. The observables indicate that model generalizes well to lattice sizes $4 \leq N \leq 128$ (interpolation), but performance drops when $N>128$ (extrapolation). Note that before for lattice sizes in the interpolation regime ($ N \leq 128$), the expected values of the observables in Figures \ref{fig:exp1_ess_M} and \ref{fig:exp1_corr_fn} display a convergence to the underlying continuum limit. }


\subsection{Multi-lattice sampling in $D=2$  dimensions}
\label{sec:exp2}
\begin{figure}[H]
    \centering
    \begin{subfigure}[t]{0.5\textwidth}
        \centering
        \includegraphics[width=.9\textwidth]{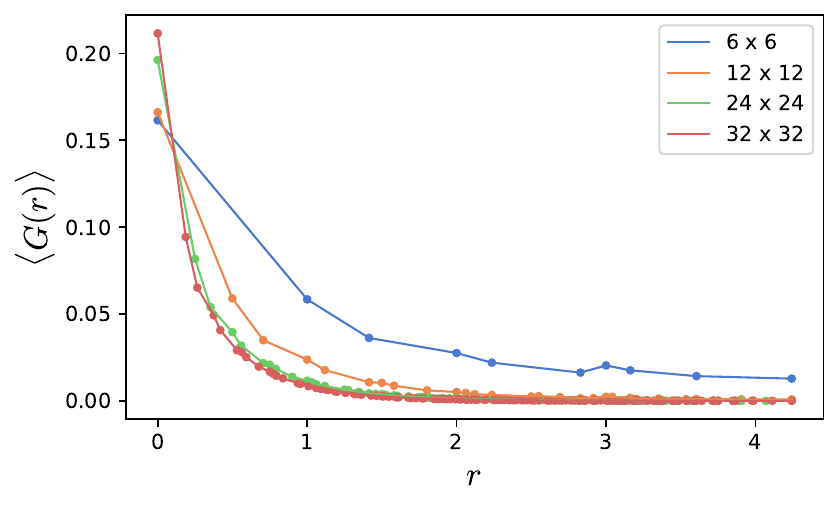}
    \end{subfigure}%
    \begin{subfigure}[t]{0.5\textwidth}
        \centering
        \includegraphics[width=.9\textwidth]{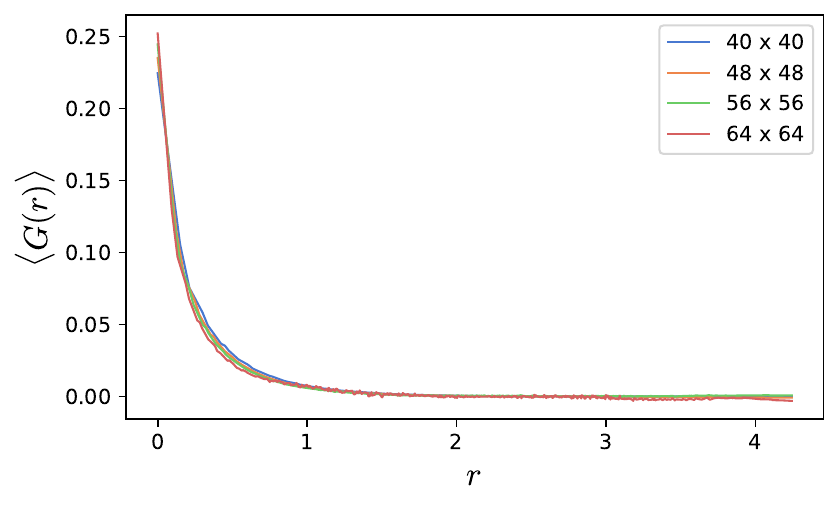}
    \end{subfigure}
    \caption{2-dimensional sampling at different lattice sizes (\S \ref{sec:exp2}). The two-point correlation function $G(x,y)$ of the second experiment computed from $16384$ samples on lattices the model was trained on (left) and on lattices the model was not trained on (right). Because of the symmetries of the task the correlation function only depends on the distance $r = |x-y|$, thus the function $G(r)$ is plotted. {Like in the previous experiment, $G(r)$ approaches the continuum limit as the model is evaluated at increasing lattice sizes.}}
    \label{fig:exp2_corr_fn}
\end{figure}

In this experiment we work with $D = 2, L = 6, m^2 = -4, g = 6.975$ (the smallest system of \cite{gerdes2022learning}). We train a model for $15000$ steps with $N=L/a$  uniformly sampled from $[6,7,8,...32]$ at each training step. We evaluate the trained model on lattices up to size $64 \times 64$. We report the effective sample size, as well as the expected value of the observables $M$ and $|M|$ (Table \ref{tab:exp2})  and the estimated correlation function at different lattice sizes (Figure \ref{fig:exp2_corr_fn}). 

\ExplSyntaxOn
\file_get:nnN {\tablepath/exp2.csv}{\ExplSyntaxOff} \tableexptwo
\file_get:nnN {\tablepath/exp3.csv}{\ExplSyntaxOff} \tableexpthree
\ExplSyntaxOff
\begin{center}

\begin{table}[H]
    \centering
    \caption{2-dimensional sampling at different lattice sizes (\S \ref{sec:exp2}). Effective sample size and expected value of the observables $M,|M|$ computed at different lattice sizes. The four rightmost columns correspond to lattice sizes the model was not trained on.}
    \label{tab:exp2}
    \begin{tabular}{c|cccccc|cccccccc}
        \toprule
        \expandafter\empty\tableexptwo
        \bottomrule
    \end{tabular}
\end{table}
\end{center}

As in the previous experiment, the model does not extrapolate well to lattices much larger than those that it was trained on. It is worth noting that performance as a function of lattice size does not drop suddenly and it is still acceptable on lattices slightly larger than the largest training lattice.  This observation motivates the following experiment.

\subsection{Faster convergence on a target lattice size by pretraining on smaller ones}
\label{sec:exp3}
\begin{figure}[H]
    \centering
    \includegraphics[width=\textwidth]{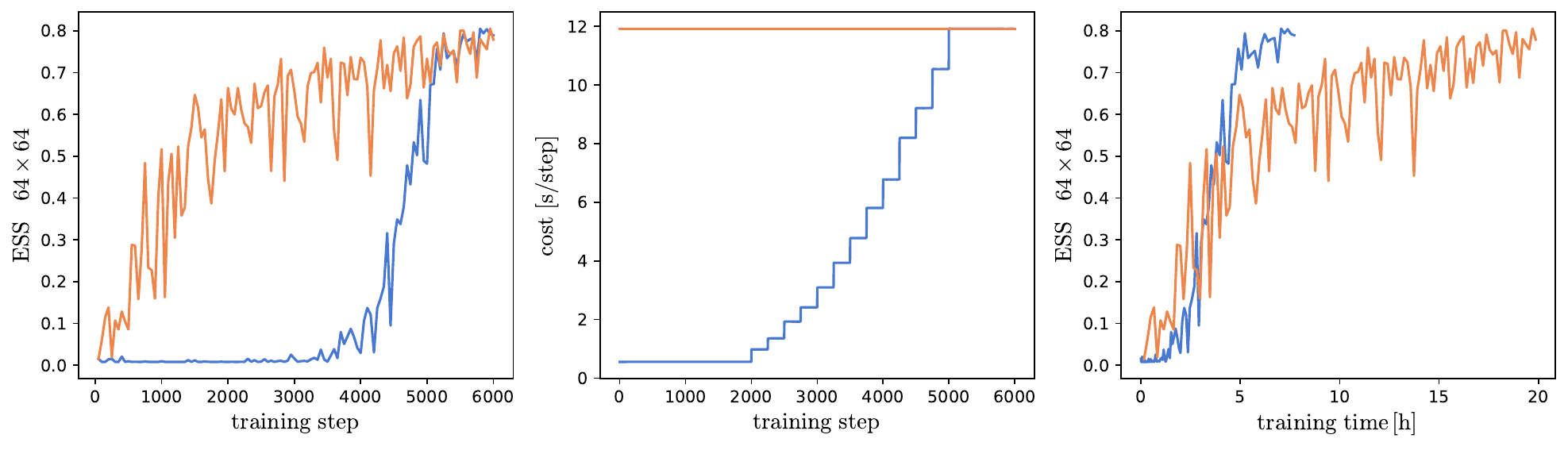}
    \caption{Pretraining on smaller lattices (\S \ref{sec:exp3}). $ESS$ estimated during training on $128$ samples  plotted against the number of training steps (left) and training time (right). Time required to take a single step (center). All plots contain two curves, one for the model that is trained on the sequence of increasing lattice sizes (blue) and one that is only trained on the $64 \times 64$ lattice (orange). We also refer the reader to
    Figure \ref{fig:exp_wandb_curves_full} in the appendix that shows the $ESS$ values on all lattice sizes during training.}
    \label{fig:exp3_compact_curves}
\end{figure}

\begin{SCfigure}[][h]
    \centering
    \caption{Pretraining on smaller lattices (\S \ref{sec:exp3}). The two-point correlation function $G(x,y)$ computed from $16384$ samples. These curves are computed from  model checkpoints saved right after taking the last training step on the given lattice size. Because of the symmetries of the task the correlation function only depends on the distance $r = |x-y|$, thus the function $G(r)$ is plotted.}
    \label{fig:exp3_correlation_fn}
    \includegraphics[width=.6\textwidth]{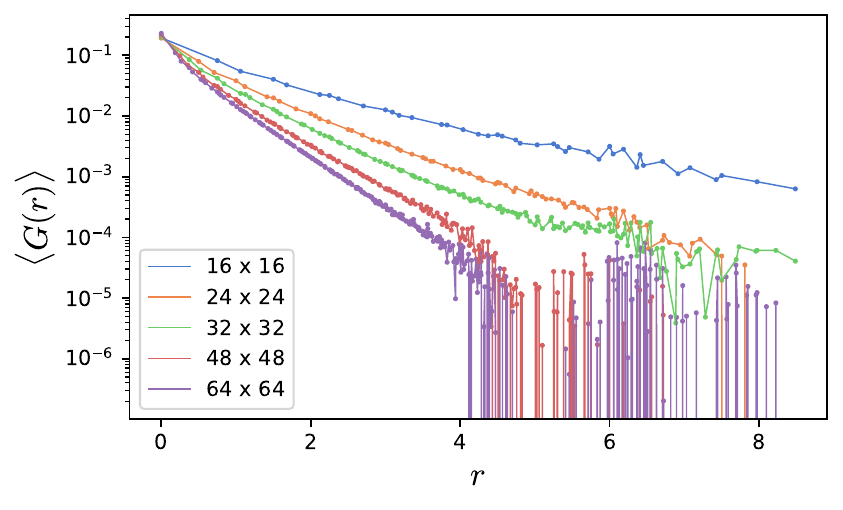}
  \end{SCfigure}

In this experiment we consider a target  with the following parameters: $D = 2, L = 12, m^2 = -4, g = 5.276, N = 64$. Instead of training directly on the $N=64$ lattice, we pretrain on a sequence of smaller lattices as they are significantly cheaper to work on. We start training on a $12 \times 12$ lattice for 2000 steps, after which we train on lattices of size $16\times 16, 20\times 20, 24 \times 24,28\times 28,32 \times 32,36 \times 36,40\times 40,44 \times 44,48 \times 48,52 \times 52,56 \times 56,60 \times 60$ for 250 training steps each. Finally, we train on the target size $64\times 64$ for 1000 steps. As a baseline, we also train the same architecture only on the target size for the same total number of steps (6000). While the performance, as measured by the effective sample size on target lattice, is comparable after training (Table \ref{tab:exp3}), the training procedure that "trained through" the smaller lattices was $\sim 2.4$-times quicker to train(Figure \ref{fig:exp3_compact_curves}).  Figure \ref{fig:exp3_correlation_fn} shows the estimated correlation function at different lattice sizes computed from  model checkpoints saved right after taking the last training step on the given lattice size.

\begin{table}[H]
    \centering
    \caption{Pretraining on smaller lattices (\S \ref{sec:exp3}). ESS values on $16384$ samples from the trained model. Since training on larger lattices degrades performance on smaller ones (Figure \ref{fig:exp_wandb_curves_full}), the model is  evaluated directly after the last training step has been performed on a given lattice size. The final column marked with $\flat$ denotes the baseline model. }
    \label{tab:exp3}
    \begin{tabular}{c|ccccc|cccc}
        \toprule
        \expandafter\empty\tableexpthree
        \bottomrule
    \end{tabular}
\end{table}

\section{Conclusion}

In this work we explored the idea of  using a neural operator-based normalizing flows for sampling from the $\phi^4$ quantum field theory. {The main advantage of this approach is that the operator-based parametrization captures a continuous representation of the task, just like a continuous field theory underlies the lattice field theory. This in particular means that a single model can be evaluated at different lattice resolutions, and can be used to approximate the continuum limit. }Experiments \ref{sec:exp1} and \ref{sec:exp2} showed that models trained on a collection of lattices do not generalize zero-shot to lattice sizes much larger than those of the training set. They do generalize with a reasonable performance to lattice sizes slightly larger than the ones it has been trained on. Making use of this observation, in experiment \ref{sec:exp3} we show that training a model on a sequence of  meshes of increasing size leads to faster training compared to training directly on the target lattice size.

\section{Acknowledgement}
The authors were supported by the Swiss National Science Foundation under grant number CRSII5\_193716 -
``Robust Deep Density Models for High-Energy Particle Physics and Solar Flare Analysis (RODEM)''.
We thank Samuel Klein for discussions.

\bibliography{main}
\newpage
\appendix             
\section*{Additional figures}
    \begin{figure}[H]
        \centering
        \includegraphics[width=.9\textwidth]{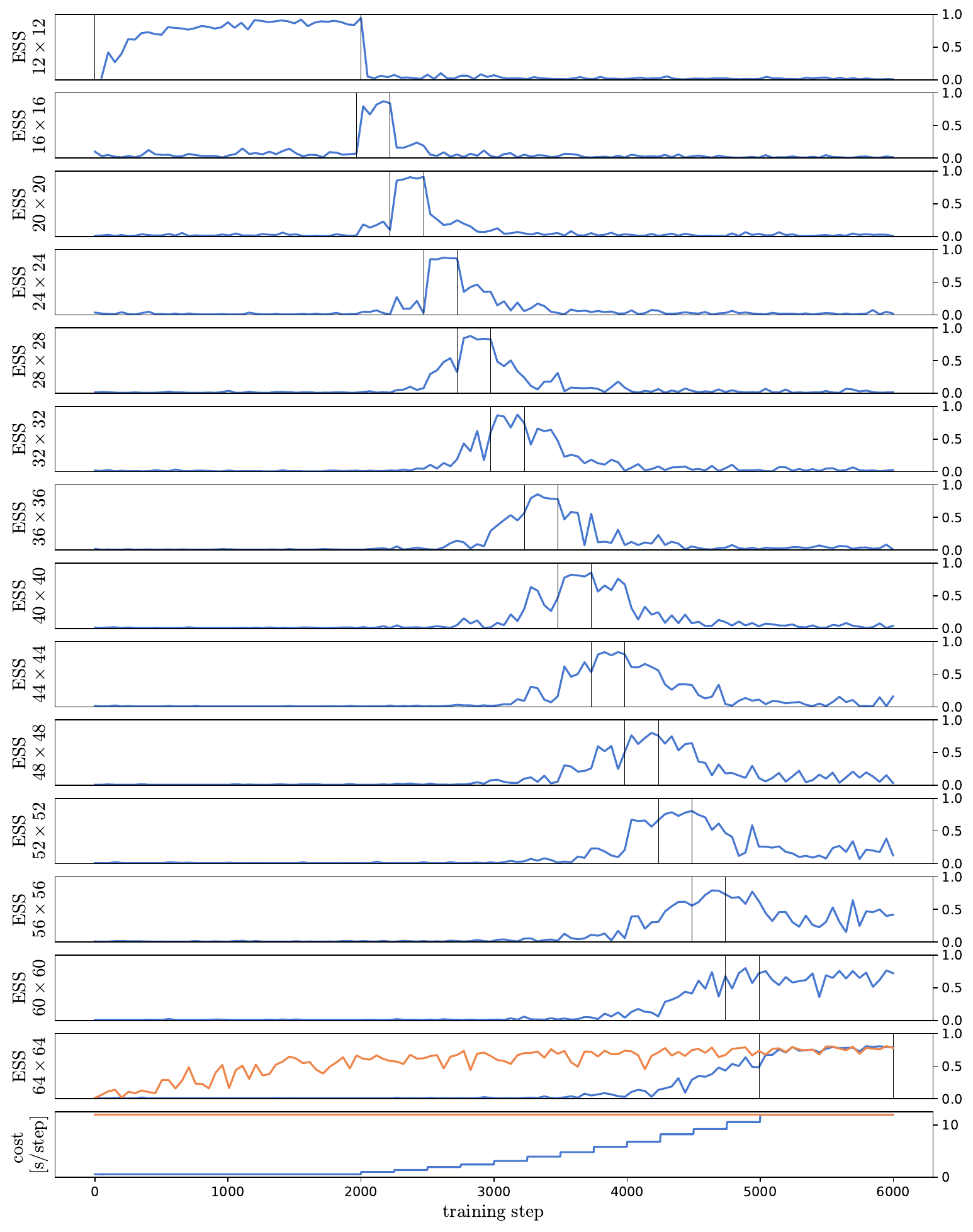}
        \caption{Experiment \ref{sec:exp3}. ESS values computed during training from 128 samples on all the lattices the sees during training. The two thin vertical lines denote the interval during which the model is trained on the given lattice size. The orange curve corresponds to the baseline model only trained on the $64 \times 64$ lattice.}
        \label{fig:exp_wandb_curves_full}
    \end{figure}

    \begin{figure}[H]
        \centering
        \includegraphics[width=.9\textwidth]{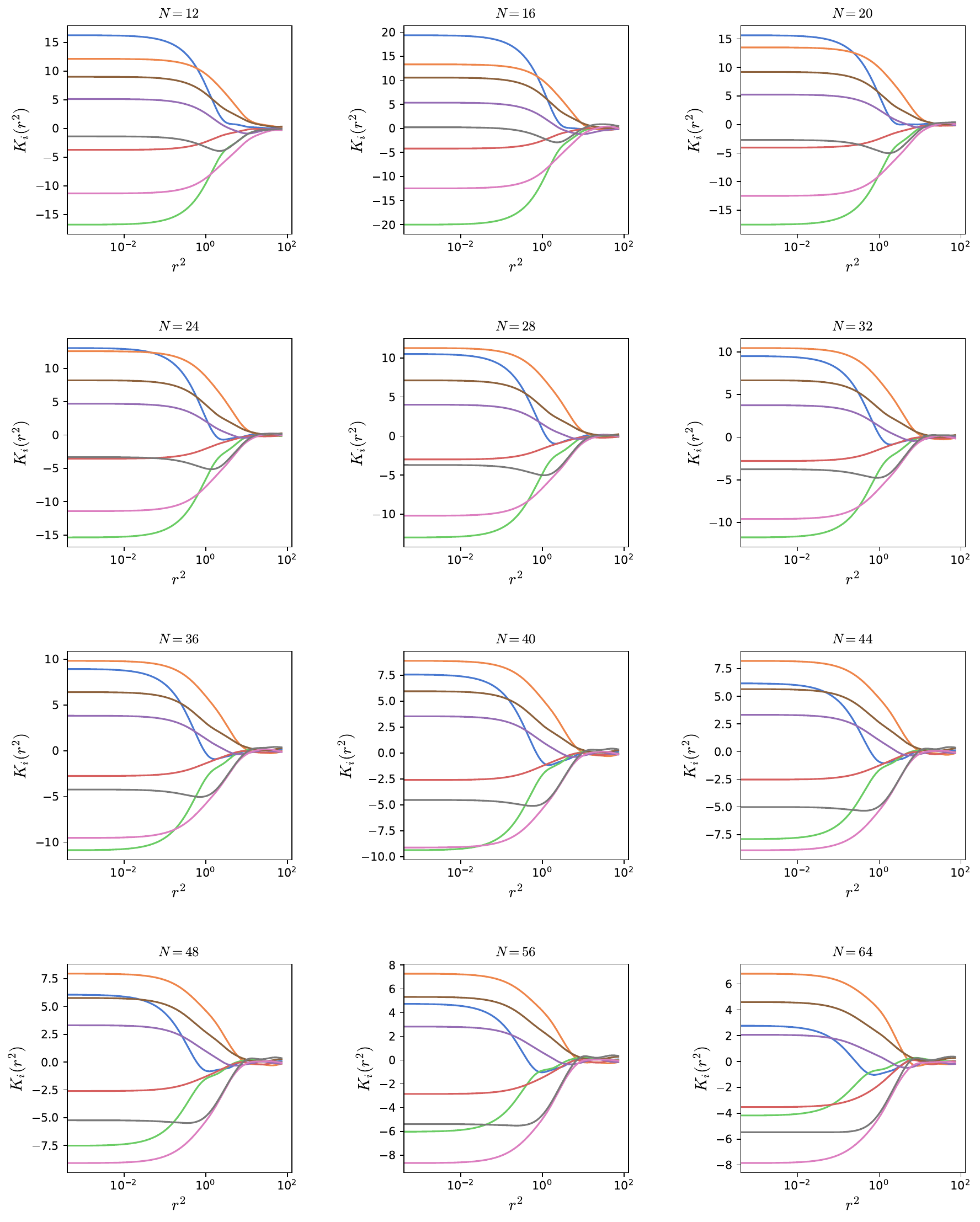}
        \caption{Experiment \ref{sec:exp3}. Kernels (Section \S \ref{sec:our_arch}) learnt by the model on various lattice sizes.}
        \label{fig:kernels}
    \end{figure}

\end{document}